\def\year{2021}\relax
\documentclass[letterpaper]{article} 
\usepackage{aaai21}  
\usepackage{times}  
\usepackage{helvet} 
\usepackage{courier}  
\usepackage[hyphens]{url}  
\usepackage{graphicx} 
\usepackage{natbib}  
\usepackage{caption} 
\usepackage[switch]{lineno}
\usepackage{xcolor}
\usepackage{amssymb}
\usepackage{amsmath}
\usepackage{dsfont}
\usepackage{latexsym}
\usepackage{graphicx}
\usepackage{accents}
\usepackage{amsthm}

\newenvironment{smalleralign}[1][\small]
 {\par\nopagebreak\leavevmode\vspace*{-\baselineskip}%
  \skip0=\abovedisplayskip
  #1%
  \def\maketag@@@##1{\hbox{\m@th\normalfont\normalsize##1}}%
  \abovedisplayskip=\skip0
  \align}
 {\endalign\ignorespacesafterend}
\makeatother

\newcommand\vartextvisiblespace[1][0.8em]{%
  \makebox[#1]{%
    \kern.04em
    \vrule height.3ex
    \hrulefill
    \vrule height.3ex
    \kern.05em
  }
}

\newcommand{\bfx}{\textbf{x}}

\title{The Struggles and Subjectivity of Feature-Based Explanations: \\ Shapley Values vs.\ Minimal Sufficient Subsets}
\author {
    Oana-Maria Camburu\textsuperscript{\rm 1, 2} \quad
    Eleonora Giunchiglia\textsuperscript{\rm 1} \quad
    Jakob Foerster\textsuperscript{\rm 3} \\
    Thomas Lukasiewicz\textsuperscript{\rm 1, 2} \quad
    Phil Blunsom\textsuperscript{\rm 1, 4}\\
}

\affiliations {
    \textsuperscript{\rm 1} University of Oxford, UK \quad \\
    \textsuperscript{\rm 2} Alan Turing Institute, London, UK \\
    \textsuperscript{\rm 3} Facebook AI Research, California, USA \quad\\
    \textsuperscript{\rm 4} DeepMind, London, UK \\
    \texttt{firstname.lastname@cs.ox.ac.uk} \quad \texttt{jnf@fb.com}\\
}

\begin{document}
\maketitle

\begin{abstract}
For neural models to garner widespread public trust and ensure fairness, we must have human-intelligible explanations for their predictions. 
Recently, an increasing number of works focus on explaining the predictions of neural models in terms of the relevance of the input features. In this work, we show that feature-based explanations pose problems even for explaining trivial models. We show that, in certain cases, there exist at least two ground-truth feature-based explanations, and that, sometimes, neither of them is enough to provide a complete view of the decision-making process of the model. Moreover, we show that two popular classes of explainers, Shapley explainers and minimal sufficient subsets explainers, target fundamentally different types of ground-truth explanations, despite the apparently implicit assumption that explainers should look for one specific feature-based explanation. These findings bring an additional dimension to consider in both developing and choosing explainers.
\end{abstract}

\section{Introduction}
A large number of explanatory methods have been developed with the goal of shedding light on black-box neural models \citep{lime, shap, saliency, integrated-grads, what-made-you-do-this, esnli, zeynep, tcav}. The majority of these methods explain the prediction of a model in terms of relevance of the input features (e.g., tokens for text). 
In this work, we show that explaining the predictions of even trivial models using only input features can be problematic. We show that there can be more than one ground-truth feature-based explanation 
and that two prevalent classes of explainers---Shapley explainers and minimal sufficient subsets (minimal sufficient subsets) explainers---target fundamentally different ground-truth explanations, without explicitly mentioning it. 
On the contrary, current works seem to imply that there is only one ground-truth feature-based explanation for a prediction. We reveal strengths and limitations of each of the two types 
and show that, sometimes, neither of them can unambiguously reflect the decision-making process of a model. 
Our findings encourage deeper reflections on the types of explanations that one aims to provide, and give users an additional dimension to consider in order to pick an explainer.

\begin{figure*}[t]
    \centering
    \centerline{\scalebox{0.9}{\begin{tabular}{ll|ll}
        \multicolumn{4}{l}{\hspace*{25ex}$m$: \textsc{if} ``very good'' \textsc{in} \textsc{input}: \textsc{return 0.9;} }\\
         \multicolumn{4}{l}{\hspace*{25.7ex}\hspace{0.4cm} \textsc{else if} ``nice'' \textsc{in input}: \textsc{return 0.7;}} \\
 \multicolumn{4}{l}{\hspace*{25.7ex}\hspace{0.4cm} \textsc{else if} ``good'' \textsc{in input}: \textsc{return 0.6;}}\\
\multicolumn{4}{l}{\hspace*{25.7ex}\hspace{0.4cm} \textsc{else return 0.}} \\  
\multicolumn{4}{l}{}\\
 \multicolumn{2}{l|}{$\textbf{x}^{1}$: ``The movie was good, it was actually nice.''} & \multicolumn{2}{l}{$\textbf{x}^{2}$: ``The movie was nice, in fact, it was very good.''}\\
 $m(\textbf{x}^{1}) = 0.7$ & & $m(\textbf{x}^{2})=0.9$ &\\ 
 & & & \\
 \underline{Shapley explanation} & \underline{MSS explanation} &  \underline{Shapley explanation} & \underline{MSS explanation}\\
 1. ``nice'': 0.4 & \{``nice''\} & 1. ``good'': 0.417 & \{``good'', ``very''\}\\
  2. ``good'': 0.3 & & 2. ``nice'': 0.367\\ 
 & & 3. ``very'': 0.116\\
    \end{tabular}}}
    \caption{Examples of cases with at least two ground-truth feature-based explanations. The Shapley values were computed via Equation~\ref{eq:shapley}; the non-mentioned features received 0 weight. MSS stands for minimal sufficient subset.} 
    \label{fig:2gt}
\end{figure*}

\section{Background}
Among existing feature-based explainers, there are two major classes: (i)~\textit{feature-additive} and (ii)~\textit{minimal sufficient subsets}. To formally describe them, let $m$ be a model to be explained and $\textbf{x}$ an instance with a potentially variable number $n = |\bfx|$ of features: $\textbf{x} = (x_1, x_2, \ldots , x_n)$. For example, $x_i$ may be the $i$-th token in the input~text~$\textbf{x}$. 

\subsection{Feature-additivity} 
A feature-additive explanation for the prediction $m(\textbf{x})$ consists of a set of importance weights $\{w_i(m, \textbf{x})\}_i$ associated with the features in $\bfx$ such that their sum approximates the prediction\footnote{For classification, the probability of the predicted class.} minus the bias of the model:
$\sum_{i=1}^{|\bfx|} w_i(m, \textbf{x})$ $=$ $m(\textbf{x}) - m(\textbf{b})$,
where $\textbf{b}$ is the baseline input, i.e., an input that brings no information, such as a zero-vector \citep{l2x}. 
The higher the absolute value of a weight, the more important the feature is for the prediction. The sign indicates whether the feature pulls towards the prediction (positive) or against it (negative).
A large number of explanatory methods are feature-additive \citep{lrp, deeplift, lime, shap, integrated-grads}. 


\vspace*{-1ex}
\paragraph{Shapley values.}
\citet{shap} aim to unify all feature-additive explanatory methods by showing that the only set of importance weights that verify three properties (local accuracy, missingness, and consistency---we refer to their paper for details) is given by the Shapley values from coalitional game theory, i.e.,
\begin{smalleralign}[\footnotesize]\label{eq:shapley}
\begin{split}
w_i(m, \textbf{x}) =\!\!\!\!\!\!\!\!
\sum_{\textbf{x}' \subseteq \textbf{x} \setminus \{x_i\}} \!\!\!\!\!\!\!
\frac{|\textbf{x}'|! (|\textbf{x}|\,{-}\,|\textbf{x}'|\,{-}\,1)!}{|\textbf{x}|!} [m(\textbf{x}' \,{\cup}\, \{x_i\}) \,{-}\, m(\textbf{x}')]\,, 
\end{split}
\end{smalleralign}%
where the sum enumerates over all subsets $\textbf{x}'$ of features in $\textbf{x}$\\

\noindent that do not include the feature for which the weight is computed. 
\citet{shap} provide methods (e.g., KernelSHAP) for approximating the Shapley values. 

\subsection{Minimal Sufficient Subsets} 
An minimal sufficient subset explanation of $m(\textbf{x})$ consists of a subset of features $\text{mss}(m, \textbf{x}) \subseteq \textbf{x}$ such that the model gives (almost) the same prediction based only on the information from $\text{mss}(m, \textbf{x})$, and no other subset of $\text{mss}(m, \textbf{x})$ leads $m$ to the same prediction, i.e.,
%
\begin{equation}\label{eq:mss1}
    m(\text{mss}(m, \textbf{x})) = m(\textbf{x}) \text{ and}
\end{equation}
\begin{equation}\label{eq:mss2}
\forall \textbf{s}' \subset \text{mss}(m, \textbf{x}) \colon m(\textbf{s}') \neq m(\textbf{x}).
\end{equation}
To compute the prediction on a subset of features, one can eliminate the other features by occlusion (via a baseline feature) or deletion (if possible).
Minimal sufficient subsets explainers are increasingly popular \citep{l2x, invase, what-made-you-do-this, anchors}.

\begin{figure*}[t]
    \centering
    \centerline{\scalebox{0.67}{\begin{tabular}{ll|ll|ll}
\multicolumn{6}{l}{\textsc{\hspace{1.3cm}1. aspect\_indicators} = \{``taste'', ``smell'', ``appearance''\} (\textsc{and any variation, such as} ``Tastes'')} \\
\multicolumn{6}{l}{\textsc{\hspace{1.3cm}2. sentiment\_indicators} = \{``amazing'' $\rightarrow$ 1, ``good'' $\rightarrow$ 0.6, ``refreshing'' $\rightarrow$ 0.6, ``bad'' $\rightarrow$ $-$0.6, ``peculiar''$\rightarrow$ $-$0.3, ``horrible'' $\rightarrow$ $-$1\}}\\
\multicolumn{6}{l}{\textsc{\hspace{1.3cm}3. A sentiment indicator is associated to its closest aspect (occluded tokens are counted).}}\\
\multicolumn{6}{l}{\textsc{\hspace{1.3cm}4. An occluded token is considered to be neutral.}}\\
\multicolumn{6}{l}{\textsc{\hspace{1.3cm}5. If more sentiment indicators are associated to an aspect, then}}\\ 
\multicolumn{6}{l}{\hspace{1.7cm}(i) \textsc{if all are of the same sign: the score for that aspect is the score of the strongest sentiment.}}\\ 
\multicolumn{6}{l}{\hspace{1.7cm}(ii) \textsc{if there are both positive and negative sentiments associated to the aspect: the score for that aspect is}}\\
\multicolumn{6}{l}{\hspace{2.3cm}\textsc{the thresholded sum of scores (max(min(sum\_scores, 1), $-$1)).}}\\ 
\multicolumn{6}{c}{}\\
\multicolumn{2}{l|}{$m^{\text{O}}$: s = \textsc{sum of scores of aspects}} & \multicolumn{2}{l|}{$m^{\text{S}}$: \textsc{return score of smell}} & \multicolumn{2}{l}{$m^{\text{T}}$: \textsc{return score of taste}} \\
\multicolumn{2}{l|}{\hspace{0.6cm} \textsc{return max(min}(s, 1), $-$1)} & \multicolumn{2}{l|}{} & \multicolumn{2}{l}{} \\
\multicolumn{2}{l|}{} & \multicolumn{2}{l|}{$\bfx^{\text{S1}}$: ``Tastes horrible, peculiar smell.''} & \multicolumn{2}{l}{$\bfx^{\text{T1}}$: ``Tastes good, refreshing.''}\\
\multicolumn{2}{l|}{$\bfx^{\text{O}}$: ``The beer has an amazing appearance,} & \multicolumn{2}{l|}{$m^{\text{S}}(\bfx^{\text{S1}})=-0.3$} & \multicolumn{2}{l}{$m^{\text{T}}(\bfx^{\text{T1}})=0.6$}\\
\multicolumn{2}{l|}{\hspace{0.8cm}a good smell, a bad taste.''} & \multicolumn{2}{l|}{} & \multicolumn{2}{l}{}\\
\multicolumn{2}{l|}{$m^{\text{O}}(\bfx^{\text{O}})=1$} & \underline{Shapley explanation}               & \underline{MSS explanation} & \underline{Shapley explanation}               & \underline{MSSs explanations}\\
\multicolumn{2}{l|}{} & 1. ''smell``: $-$0.29 & \{``peculiar'', ``smell''\} & 1. ``Tastes'': 0.4  & \{``Tastes'', ``good''\},\\
\underline{Shapley explanation}               & \underline{MSS explanation}      &      2. ''Tastes``: 0.26          &         &        2. ``good'': 0.1,   & \{``Tastes'', ``refreshing''\}  \\
1. ``amazing'': 0.52 & \{``amazing'', ``appearance''\} & 3. ''horrible``: $-$0.14 &  & \hspace{0.43cm}``refreshing'': 0.1  &   \\
2. ``good'': 0.40 &  & 4. ''peculiar``: $-$0.13  & &  &\\
3. ``bad'': $-$0.23 &  &  &  &  & \\
4. ``smell'': 0.15 & &&  & & \\
5. ``appearance'': 0.12 &  & \multicolumn{2}{l|}{$\bfx^{\text{S2}}$: ``Tastes amazing, peculiar smell.''}  &  \multicolumn{2}{l}{$\bfx^{\text{T2}}$: ``Tastes amazing. The smell is also amazing.''} \\
6. ``taste'': 0.03 &  &  $m^{\text{S}}(\bfx^{\text{S2}})=-0.3$ & &  $m^{\text{T}}(\bfx^{\text{T2}})=0.6$ & \\
\multicolumn{2}{l|}{} & \multicolumn{2}{l|}{}  & \multicolumn{2}{l}{} \\
\multicolumn{2}{l|}{} &  \underline{Shapley explanation}               & \underline{MSS explanation}  &\underline{Shapley explanation}               & \underline{MSSs explanations} \\
&  & 1. ``peculiar'': $-$0.27 & \{``peculiar'', ``smell''\} &  1. ``Tastes'': 0.58  & \{``Tastes'', ``$\text{amazing}^1$''\}, \\
& & 2. ``smell'': $-$0.10 &  & 2. ``$\text{amazing}^1$'': 0.42   & \{``Tastes'', ``$\text{amazing}^2$''\}     \\
&  & 3. ``amazing'': 0.05 & &  3. ``$\text{amazing}^2$'': 0.08  &  \\
& & 4. ``Tastes'': 0.02 & & 4. ``smell'': $-$0.08 & \\
\end{tabular}}}
\caption{Examples illustrating the strengths and limitations of the two types of feature-based explanations, as presented in Section~\ref{sec:s-lim}. The five rules are common to all three models. The Shapley values were computed via Equation~\ref{eq:shapley}, and written in decreasing order of their importance (absolute value); the non-mentioned features received 0 weight. In the last example, the superscript of ``amazing'' differentiates between its two occurrences. MSS stands for minimal sufficient subset.}
\label{fig:stren_lim}
\end{figure*}

\section{Two Types of Ground-Truth Feature-Based Explanations}
We show that, in certain cases, there exist more than one ground-truth feature-based explanation for a prediction, and that Shapley and minimal sufficient subsets explainers target two such different types of ground-truth explanations for even trivial models. We also reveal strengths and limitations for each type, and we show that, sometimes, none of them is enough to provide a complete view on a model.
To illustrate our findings, we give examples of hypothetical sentiment analysis models.  
Such models take as input a review and output a score reflecting the sentiment of the review towards the object of interest or towards an aspect of the object (for multi-aspect reviews) \citep{beer-annot}. We treat 
the scores as real numbers linearly reflecting the intensity of the sentiment, with $-$1 the most negative, 1 the most positive, hence, 0 being the neutral score. %
Without loss of generality, we assume that scores that differ by at least 0.1 indicate significantly different sentiments. 

To illustrate the existence of more than one ground-truth feature-based explanation for a prediction of a model on an instance, in Fig.~\ref{fig:2gt}, we present an example of a hypothetical sentiment analysis regression model $m$.
The prediction of $m$ on the instance $\textbf{x}^{1}$ is $m(\textbf{x}^{1}) = 0.7$, as ``nice'' is present in the instance. Hence, one can argue that ``nice'' is the only important feature for this prediction, while ``good'' would be the only important feature for a different prediction (of 0.6). 
On the other hand, one may argue that ``good'' should also be flagged as important, because if ``nice'' is eliminated, then the model relies on ``good'' to provide a score as high as 0.6 instead of the much lower default of 0. 
The difficulty faced when trying to explain $m$ with feature-based explanations is even more pronounced on the instance $\textbf{x}^{2}$. The model predicts $m(\textbf{x}^{2}) = 0.9$, as ``very good'' is in $\textbf{x}^{2}$. Hence, an explanation that states that the features ``very'' and ``good'' are the only important features for this prediction is one ground-truth explanation. However, if ``good'' is eliminated from this instance, the model relies on ``nice'' (and not on ``very'') to provide a score as high as 0.7, while if both ``good'' and ``nice'' are eliminated, then the score drops all the way to 0. From this perspective, ``nice'' can be seen as more important than ``very'', and an explanation that ranks ``good'', ``nice'', and ``very'' in this order of importance is also a ground-truth explanation. Therefore, there are two fundamentally different perspectives on what a ground-truth feature-based explanation should be, even for this trivial model. 

\vspace*{-1ex}
\paragraph{Shapley explanations vs.\ Minimal sufficient subsets explanations.}
The above two types of ground-truth explanations are separately advocated by Shapley and minimal sufficient subsets explainers, as shown in Fig.~\ref{fig:2gt}. In particular, notice how the Shapley explanation attributes to ``very'' about three times less importance than to ``nice'' for $m(\textbf{x}^{2})$, while ``nice'' does not even make it in the minimal sufficient subset explanation. 

This difference stems from the fact that Shapley values were introduced to promote \textit{fairness} in distributing a total gain among the players of a coalition \citep{shapley1}. Hence, they provide the \textit{average} importance of the features on a \textit{neighborhood} of the instance, taking into account each player's performance in any sub-coalition. On the other hand, minimal sufficient subsets explanations provide the features that are \textit{pointwise} important for the prediction on the instance \textit{in isolation}, rewarding only the players that are crucial inside the \textit{full} coalition. 

\vspace*{-1ex}
\paragraph{Not always distinct.} Although the two types of explanations are distinct for certain cases, in other cases they coincide. For example, for the model $m$ in Fig.~\ref{fig:2gt} and instances that contain exactly one of the subphrases ``very good'', ``nice'', and ``good'', such as ``The movie was good.'', both types of explanations point towards the same features. This has likely hindered the fact that the two types of explanations are fundamentally distinct.

\vspace*{-1ex}
\paragraph{Literature.}
To our knowledge, the existence of two types of ground-truth feature-based explanations was only briefly alluded by \citet{integrated-grads} when explaining the function $\min(x_1, x_2)$ on the instance $x_1 = 1, x_2 = 3$. Their method attributes the whole importance weight (of $1 = \min(1, 3) - \min(0, 0)$) to the \textit{critical} feature $x_1$, while Shapley attributes $0.5$ importance weight to each feature. They mention that preferring one explanation over the other is subjective. 

On the other hand, \citet{shap} state\footnote{Via a graph of results (their Fig.~4).} that all participants in their user study explained the function $\max(x_1, x_2, x_3)$\footnote{Devised as a story of three people making money based on the maximum score that any of them achieved.} on the input $x_1 \,{=}\, 5, x_2 \,{=}\,4, x_3\,{=}\, 0$ by attributing the importance weights of $3$ for $x_1$, 2 for $x_2$, and 0 for $x_3$, implying that the Shapley explanation is the only ground-truth explanation for this function. However, the authors do not mention the number of participants nor the guidelines that they received.  

Moreover, current works, such as \citep{l2x} and \citep{invase}, compare minimal sufficient subsets explainers (e.g., L2X and INVASE) with Shapley explainers on identifying the features used by a model trained on synthetic datasets, where the ground-truth important features are known. While the particular synthetic datasets used in these works do not violate either of the two ground-truth explanations, this is not mentioned at the time of comparison. Such comparisons risk inducing the idea that there is always only one feature-based explanation. 

\subsection{Strengths and Limitations}
\label{sec:s-lim}
In this section, we reveal certain strengths and limitations of the two approaches presented above.

\vspace*{-1ex}
\paragraph{Redundant features.}
By looking only at the Shapley explanation for $m(\bfx^{1})$ in Fig.~\ref{fig:2gt}, one cannot know whether (1)~the model requires \textit{both} features ``nice'' and ``good'' to make its prediction of $0.7$ (which is not the case for $m$), or (2)~one of these features is redundant in presence of the other (which is the case for $m$). In contrast, minimal sufficient subsets explanations do not contain redundant features (Equation~\ref{eq:mss2}), and hence, the minimal sufficient subset explanation for $m(\bfx^{1})$ is able to distinguish between the two scenarios. 

\vspace*{-1ex}
\paragraph{Feature cancellations: genuine vs.\ artefacts.}
In certain cases, there exist features that cancel each other out. 
Consider the model $m^{\text{O}}$ in Fig.~\ref{fig:stren_lim}, which infers the overall sentiment on a beer from a multi-aspect review by adding up the scores that it associates to each aspect in the review. 
On the instance $\bfx^{\text{O}}$, $m^{\text{O}}$ predicts $1$ by taking into account all three aspects. However, the minimal sufficient subset explanation is \{``amazing'', ``appearance''\}
---it does not contain the features ``bad'', ``taste'', ``good'', and ``smell'', due to Equation~\ref{eq:mss2}. 
Arguably, users may want to see the features from such a \textit{genuine cancellation} in the explanation. 
Note that these features are flagged as important by the Shapley explanation, which, nonetheless, does not clearly indicate the perfect cancellation between ``good smell'' and ``bad taste''. Moreover, the Shapley explanation gives the impression that ``smell'' and ``appearance'' are much more important (0.15 and 0.12) than ``taste'' (0.03), when, by design, $m^{\text{O}}$ equally takes into account all aspects. Hence, neither the Shapley nor the minimal sufficient subset explanation is well reflecting the decision-making process for $m^{\text{O}}(\bfx^{\text{O}})$.

Artefacts may occur when eliminating features from an instance, distorting the importance of certain features. Model $m^{\text{S}}$ in Fig.~\ref{fig:stren_lim} illustrates such an example.
When $m^{\text{S}}$ is applied to $\bfx^{\text{S1}}$, it predicts $-$0.3, and the minimal sufficient subset explanation is \{``peculiar'', ``smell''\}, which, arguably, best reflects the decision-making process for $m^{\text{S}}(\bfx^{\text{S1}})$.  
However, in the Shapley explanation ,``Tastes'' appears to be twice more important than ``peculiar'', and ``horrible'' appears as important as ``peculiar'', even though ``peculiar'' is the actual sentiment indicator for smell. 
Furthermore, note how the Shapley importance weights dramatically change when only the sentiment on taste is changed in instance $\bfx^{\text{S2}}$, even though $m^{\text{S}}$ does not rely on the sentiment on taste to predict the sentiment on smell.  

\paragraph{Multiple minimal sufficient subsets: genuine vs.\ artefacts.}
In certain cases, there can exist multiple minimal sufficient subsets explanations for one prediction.
For example, for the model $m^{\text{T}}$ and instance $\bfx^{\text{T1}}$ in Fig.~\ref{fig:stren_lim}, either of the features ``good'' and ``refreshing'' leads to the score of 0.6. Ideally, minimal sufficient subsets explainers provide all the genuine minimal sufficient subsets, e.g., both \{``Tastes'', ``good''\} and \{``Tastes'', ``refreshing''\}. However, many minimal sufficient subsets explainers are designed to retrieve only one minimal sufficient subset \citep{l2x, invase}. An exception is the SIS explainer \citep{what-made-you-do-this}, which retrieves a set of disjoint minimal sufficient subsets, which might also not be exhaustive (e.g., SIS would not retrieve the second minimal sufficient subset explanation for $m^{\text{T}}(\bfx^{\text{T1}})$, because ``Tastes'' is already taken by the first minimal sufficient subset).
On the other hand, the Shapley explanation gives the same importance to both ``good'' and ``refreshing''. However, with this explanation alone, one would not be able to know whether both ``good'' and ``refreshing'' are necessary to be present or if each individually suffices for the prediction of $0.6$.

Artefacts occurring when eliminating features can also make certain subsets of features appear as minimal sufficient subsets. 
For example, for $m^{\text{T}}$ and $\bfx^{\text{T2}}$ in Fig.~\ref{fig:stren_lim}, either of the two occurrences of ``amazing'' forms an minimal sufficient subset, but the second one is not reflecting the decision-making process of the model. The Shapley explanation makes the distinction in this case.

\vspace*{-1ex}
\paragraph{Discussion.}
Our findings also show that the use of both the minimal sufficient subset and the Shapley explanation together might lead to a better understanding of the decision-making process of a model. For example, for $\bfx^1$ in Fig. \ref{fig:2gt}, if one has both the Shapley and the minimal sufficient subset explanation, one would conclude that ``nice'' is enough for the prediction of the model $m$ on $\bfx^1$, and that ``good'' is redundant in presence of ``nice'' but still an informative feature for this model in general. Similarly, for the instance $\bfx^O$ in Fig. \ref{fig:stren_lim}, having both explanations allows one to infer that the model cancels out ``good smell'' and ``bad taste'', which would not have been possible to infer from any of the two explanations alone. Hence, our work also opens the path for future investigation into the ways and benefits of combining different types of explanations. 

We use simple rule-based models to expose the characteristics of the explainers, first, because one does not know how a complex neural network makes its decisions, which is the reason in the first place to develop explainers. Hence, we assess the explainers on rule-based models whose inner workings we know. 
Second, if the explainers have certain characteristics on simple models, then these characteristics are likely to persist in the more complex neural models to be explained. This is because the rule-based models that we exemplify are not at all unrealistic, as neural networks were shown to learn to sometimes rely on simple combinations of input features to provide their predictions \citep{artifacts, anchors}. 

\section{Summary and Outlook}
In this work, we showed that, for certain cases, there is more than one ground-truth feature-based explanation for the prediction of a model, and that Shapley explainers and minimal sufficient subsets explainers aim to provide two such fundamentally different ground-truth explanations. We provided insights into the strengths and limitations of these types of explanations. 

As one type of feature-based explanation may be preferred over the other in different real-world applications, the choice is best left in the hands of the users, who need to be informed of these differences in order to pick and use the explainers accordingly. Besides users, researchers also benefit from knowing the specifications of the target explanations of any explanatory method, e.g., to avoid unfairly comparing methods that aim for different explanations on the basis of one explanation being considered the only ground-truth. Therefore, this work aims to urge the community to rigorously and directly state the specifications of the target explanations of any explanatory method.  

Future work includes user studies to decide to what extent users benefit from each of these types of explanations or their combination. 
Most importantly, this work encourages further reflection on the types of explanations that we, as a community, aim to provide. 

\paragraph{Acknowledgments.}
During this work, Oana-Maria Camburu and Thomas Lukasiewicz were supported by a JP Morgan PhD Fellowship, the Alan Turing Institute under the EPSRC grant EP/N510129/1, the AXA Research Fund, the ESRC grant ``Unlocking the Potential of AI for Law'', and the EU Horizon 2020 grant 952215.
Eleonora Giunchiglia was  supported by
the EPSRC  grant EP/N509711/1 and by an Oxford-DeepMind Graduate Scholarship. 

\bibliography{aaai2021}

\begin{thebibliography}{16}
\providecommand{\natexlab}[1]{#1}
\providecommand{\url}[1]{\texttt{#1}}
\providecommand{\urlprefix}{URL }
\expandafter\ifx\csname urlstyle\endcsname\relax
  \providecommand{\doi}[1]{doi:\discretionary{}{}{}#1}\else
  \providecommand{\doi}{doi:\discretionary{}{}{}\begingroup
  \urlstyle{rm}\Url}\fi

\bibitem[{Arras et~al.(2017)Arras, Montavon, M{\"u}ller, and Samek}]{lrp}
Arras, L.; Montavon, G.; M{\"u}ller, K.-R.; and Samek, W. 2017.
\newblock Explaining Recurrent Neural Network Predictions in Sentiment
  Analysis.
\newblock In \emph{Proceedings of the 8th Workshop on Computational Approaches
  to Subjectivity, Sentiment and Social Media Analysis}, 159--168.

\bibitem[{Camburu et~al.(2018)Camburu, Rockt{\"{a}}schel, Lukasiewicz, and
  Blunsom}]{esnli}
Camburu, O.; Rockt{\"{a}}schel, T.; Lukasiewicz, T.; and Blunsom, P. 2018.
\newblock {e-SNLI: N}atural Language Inference with Natural Language
  Explanations.
\newblock In \emph{Advances in Neural Information Processing Systems 31: Annual
  Conference on Neural Information Processing Systems (NeurIPS)}, 9560--9572.

\bibitem[{Carter et~al.(2019)Carter, Mueller, Jain, and
  Gifford}]{what-made-you-do-this}
Carter, B.; Mueller, J.; Jain, S.; and Gifford, D.~K. 2019.
\newblock What made you do this? {U}nderstanding black-box decisions with
  sufficient input subsets.
\newblock In \emph{Proceedings of the International Conference on Artificial
  Intelligence and Statistics (AISTATS)}.

\bibitem[{Chen et~al.(2018)Chen, Song, Wainwright, and Jordan}]{l2x}
Chen, J.; Song, L.; Wainwright, M.; and Jordan, M. 2018.
\newblock Learning to Explain: An Information-Theoretic Perspective on Model
  Interpretation.
\newblock In \emph{Proceedings of the 35th International Conference on Machine
  Learning}, volume~80 of \emph{Proceedings of Machine Learning Research},
  883--892.

\bibitem[{Gururangan et~al.(2018)Gururangan, Swayamdipta, Levy, Schwartz,
  Bowman, and Smith}]{artifacts}
Gururangan, S.; Swayamdipta, S.; Levy, O.; Schwartz, R.; Bowman, S.; and Smith,
  N.~A. 2018.
\newblock Annotation Artifacts in Natural Language Inference Data.
\newblock In \emph{Proceedings of the 2018 Conference of the North American
  Chapter of the Association for Computational Linguistics: Human Language
  Technologies, Volume 2 (Short Papers)}, 107--112.

\bibitem[{Kim et~al.(2018)Kim, Wattenberg, Gilmer, Cai, Wexler, Viegas, and
  Sayres}]{tcav}
Kim, B.; Wattenberg, M.; Gilmer, J.; Cai, C.; Wexler, J.; Viegas, F.; and
  Sayres, R. 2018.
\newblock Interpretability Beyond Feature Attribution: Quantitative Testing
  with Concept Activation Vectors ({TCAV}).
\newblock In \emph{Proceedings of the 34th International Conference on Machine
  Learning, {ICML} 2018}.

\bibitem[{Lundberg and Lee(2017)}]{shap}
Lundberg, S.~M.; and Lee, S.-I. 2017.
\newblock A Unified Approach to Interpreting Model Predictions.
\newblock In \emph{Advances in Neural Information Processing Systems 30},
  4765--4774.

\bibitem[{McAuley, Leskovec, and Jurafsky(2012)}]{beer-annot}
McAuley, J.~J.; Leskovec, J.; and Jurafsky, D. 2012.
\newblock Learning Attitudes and Attributes from Multi-aspect Reviews.
\newblock In \emph{Proceedings of the 12th {IEEE} International Conference on
  Data Mining ({ICDM})}, 1020--1025.

\bibitem[{Park et~al.(2018)Park, Hendricks, Akata, Rohrbach, Schiele, Darrell,
  and Rohrbach}]{zeynep}
Park, D.~H.; Hendricks, L.~A.; Akata, Z.; Rohrbach, A.; Schiele, B.; Darrell,
  T.; and Rohrbach, M. 2018.
\newblock Multimodal Explanations: Justifying Decisions and Pointing to the
  Evidence.
\newblock In \emph{Proceedings of the IEEE Conference on Computer Vision and
  Pattern Recognition (CVPR)}.

\bibitem[{Ribeiro, Singh, and Guestrin(2016)}]{lime}
Ribeiro, M.~T.; Singh, S.; and Guestrin, C. 2016.
\newblock {``Why should I trust you?'': Explaining the predictions of any
  classifier}.
\newblock In \emph{Proceedings of the 22nd {ACM} {SIGKDD} International
  Conference on Knowledge Discovery and Data Mining}, 1135--1144.

\bibitem[{Ribeiro, Singh, and Guestrin(2018)}]{anchors}
Ribeiro, M.~T.; Singh, S.; and Guestrin, C. 2018.
\newblock Anchors: High-Precision Model-Agnostic Explanations.
\newblock In \emph{Proceedings of the 32nd {AAAI} Conference on Artificial
  Intelligence, (AAAI)}, 1527--1535.

\bibitem[{Shapley(1951)}]{shapley1}
Shapley, L.~S. 1951.
\newblock Notes on the n-Person Game -- {II}: The Value of an n-Person Game.
\newblock \emph{Santa Monica, Calif.: RAND Corporation.} .

\bibitem[{Shrikumar, Greenside, and Kundaje(2017)}]{deeplift}
Shrikumar, A.; Greenside, P.; and Kundaje, A. 2017.
\newblock Learning Important Features Through Propagating Activation
  Differences.
\newblock In \emph{Proceedings of the 34th International Conference on Machine
  Learning ({ICML} 2017)}, 3145--3153.

\bibitem[{Simonyan, Vedaldi, and Zisserman(2014)}]{saliency}
Simonyan, K.; Vedaldi, A.; and Zisserman, A. 2014.
\newblock Deep Inside Convolutional Networks: Visualising Image Classification
  Models and Saliency Maps.
\newblock In \emph{2nd International Conference on Learning Representations,
  {ICLR} 2014, Workshop Track Proceedings}.

\bibitem[{Sundararajan, Taly, and Yan(2017)}]{integrated-grads}
Sundararajan, M.; Taly, A.; and Yan, Q. 2017.
\newblock Axiomatic Attribution for Deep Networks.
\newblock In \emph{Proceedings of the 34th International Conference on Machine
  Learning ({ICML})}, 3319--3328.

\bibitem[{Yoon, Jordon, and van~der Schaar(2019)}]{invase}
Yoon, J.; Jordon, J.; and van~der Schaar, M. 2019.
\newblock {INVASE: I}nstance-wise Variable Selection using Neural Networks.
\newblock In \emph{Proceedings of the 7th International Conference on Learning
  Representations ({ICLR} 2019)}.

\end{thebibliography}

\end{document}